\documentclass[runningheads]{llncs}
\usepackage{graphicx}
\usepackage{tikz}
\usepackage{comment}
\usepackage{amsmath,amssymb} % define this before the line numbering.
\usepackage{color}
\usepackage{tabularx}
\usepackage{algorithm}
\usepackage{algpseudocode}
\usepackage{multicol}
\usepackage{subcaption}

\usepackage[width=122mm,left=12mm,paperwidth=146mm,height=193mm,top=12mm,paperheight=217mm]{geometry}

\begin{document}
% \renewcommand\thelinenumber{\color[rgb]{0.2,0.5,0.8}\normalfont\sffamily\scriptsize\arabic{linenumber}\color[rgb]{0,0,0}}
% \renewcommand\makeLineNumber {\hss\thelinenumber\ \hspace{6mm} \rlap{\hskip\textwidth\ \hspace{6.5mm}\thelinenumber}}
% \linenumbers
\pagestyle{headings}
\mainmatter
\def\ECCVSubNumber{9}  % Insert your submission number here

\title{Black-Box Face Recovery from Identity Features} % Replace with your title

% INITIAL SUBMISSION 
\begin{comment}
\titlerunning{ECCV-20 submission ID \ECCVSubNumber} 
\authorrunning{ECCV-20 submission ID \ECCVSubNumber} 
\author{Anonymous ECCV submission}
\institute{Paper ID \ECCVSubNumber}
\end{comment}
%******************

% CAMERA READY SUBMISSION
%\begin{comment}
\titlerunning{Black-Box Face Recovery from Identity Features}
% If the paper title is too long for the running head, you can set
% an abbreviated paper title here
%
\author{Anton Razzhigaev\inst{1,2} \and
Klim Kireev\inst{1,2} \and
Edgar Kaziakhmedov\inst{1,2} \and
Nurislam Tursynbek\inst{1,2} \and
Aleksandr Petiushko\inst{2,3}}
\authorrunning{A. Razzhigaev et al.}
% First names are abbreviated in the running head.
% If there are more than two authors, 'et al.' is used.
%
\institute{Skolkovo Institute of Science and Technology \and
Huawei Moscow Research Center \and
Lomonosov Moscow State University}
%\end{comment}
%******************
\maketitle

\begin{abstract}
In this work, we present a novel algorithm based on an iterative sampling of random Gaussian blobs for black-box face recovery, given only an output feature vector of deep face recognition systems. We attack the state-of-the-art face recognition system (ArcFace) to test our algorithm. Another network with different architecture (FaceNet) is used as an independent critic showing that the target person can be identified with the reconstructed image even with no access to the attacked model. Furthermore, our algorithm requires a significantly less number of queries compared to the state-of-the-art solution.
\end{abstract}

%%%%%%%%% BODY TEXT
\section{Introduction}

The most common characteristic to identify a person from a still image is its face. Automatic face identification is an important computer vision task with real-world applications in smartphone cameras, video surveillance systems, human-computer interaction. Following rapid progress in image classification \cite{krizhevsky2012imagenet,he2016deep}, object detection \cite{redmon2018yolov3,ren2015faster}, semantic and instance segmentation \cite{chen2018encoder,he2017mask}, Deep Neural Networks demonstrated state-of-the-art performance in face identity recognition, even in extremely challenging scenarios with millions of identities \cite{kemelmacher2016megaface}. 

Although end-to-end solutions exist, leading face recognition systems usually require a few-step procedure. First, the face is detected in the given image, and the alignment process is done. Then, the aligned face is fed to a face identification network, which converts it to descriptive feature vectors of the lower dimensionality. It is challenging to allocate those representations so different images of the same person are mapped to be closer to each other than to those of different.

Recent solutions incorporate different types of margins to the training loss to enhance the discriminative power. Current state-of-the-art publicly available model is ArcFace \cite{deng2019arcface}, a geometric method, that uses Additive Angular Margin Loss, to produce highly distinguishable features and stabilize training process. 

Besides strong performance of face recognition models in the real world, it is crucial to study and overcome their vulnerabilities, since adversaries might harm security and privacy aspects of such systems. Face recognition systems might be maliciously attacked from different perspectives. Impersonation and dodging attacks aim to fool the network, by wearing specifically designed accessories such as glasses \cite{sharif2016accessorize}. 2D and 3D spoofing attacks have been demonstrated in practical applications of face identification systems such as face unlock systems \cite{liu20163d,patel2016secure}. 

\begin{figure*}
\begin{center}
   \includegraphics[width=\linewidth]{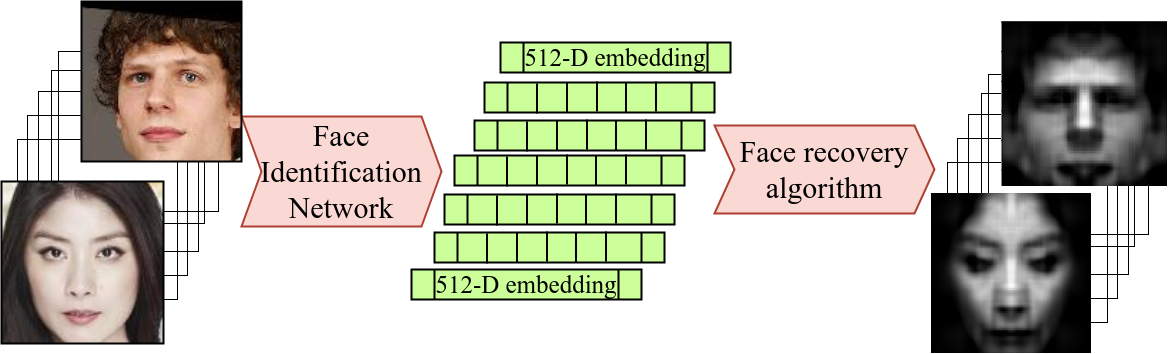}
\end{center}
    \caption{The schematic of the face recovery procedure from the identity features.}
%   \caption{The overall scheme of the reconstruction system. The faces are fed to face identification models and obtained embeddings are stored. Then face recovery algorithm reconstructs original faces given only embeddings.}
\label{example}
\end{figure*}

Another critical vulnerability of a face recognition system is the leakage of data, as face embeddings (face identity features) might be reconstructed into recognizable faces. In this paper, we consider black-box scenario (see Fig. \ref{example}), i.e. we only receive embedding produced by face identification model for our requesting image, without access to the model's architecture, since unknown embeddings might be exposed or hacked, and using corresponding target face recognition APIs we can request necessary output.

%\subsection{Adversarial attacks and Face Reconstruction}

%Adversarial attacks in the physical domain on face recognition models were the initial inspiration for this work. Following the recent research on real-world face adversarial attack \cite{komkov2019advhat}, we noticed that adversarial patches have noticeable facial semantic features, like eyebrows, eyes or forehead (Fig.~\ref{adversarial}). So the question arose if it is possible to generate the whole target face with an adversarial example technique: maximizing face-id system similarity. A typical adversarial example is a small perturbation of the initial image, which contains custom-crafted high-frequency noise rather than semantic features. To move from adversarial high-frequency noise to facial images, we developed the novel optimization procedure restricting the high-frequency adversarial behavior.

% \begin{figure}[h]
% \begin{center}
%   \includegraphics[width=0.3\linewidth]{me.jpg}
% \end{center}
%   \caption{The generated example of a physical domain adversarial patch with semantic patterns (like eyes and eyebrows).}
% \label{adversarial}
% \end{figure}

\textbf{Main contributions of this work are the following:}
\begin{itemize}
    \item We proposed the novel face recovery method in the black-box setup;
    \item We quantitatively and qualitatively demonstrated the superiority of our method compared to the previous one;
    \item The proposed method works without a prior knowledge such as a training dataset from the same domain;
    \item We evaluated our method and its competitor with an independent critic;
    \item We proved that the result is the same for the train and test datasets.
\end{itemize}

\begin{table*}
\begin{center}
\caption{Comparison table}
\label{comp_table}
% \begin{adjustbox}{width=\textwidth,center}
\begin{tabularx}{0.915\textwidth}{|l|c|c|c|}
\hline
Algorithm & Target model & Setting & Dataset-free \\ \hline\hline
Ours & Arcface output & Black-Box & + \\ \hline
NBNet\cite{mai2018reconstruction} & FaceNet output & Black-Box & -\\ \hline
Cole et. al. \cite{cole2017synthesizing} & FaceNet intermediate features & White-box & -  \\ \hline
CNN\cite{zhmoginov2016inverting} & FaceNet output features & White-box & - \\ \hline
Gradient wrt input \cite{mahendran2015understanding} & Any classifier output & White-box & + \\ \hline
\end{tabularx}
% \end{adjustbox
\end{center}
\end{table*}

\section{Related Work}
The remarkable progress of deep neural networks in many areas attained significant attention in the scientific community to the nature of its internal representations. Researchers always questioned how to interpret the decisions of deep learning models. One way of interpretation of neural networks in the task of pattern recognition was found to be the inverting of class outputs or hidden feature vectors. Using image pixel gradients, optimization-based inversion of image classification neural networks presented interesting results in \cite{mordvintsev2015inceptionism,erhan2009visualizing,yosinski2015understanding,mahendran2015understanding,simonyan2013deep}. The basic idea of this branch of works was to use white-box back-propagation to get input gradients and minimize the loss between the network output and the desired class to invert. Usually, due to the high dimensionality of the input and high-frequency gradients, it is required to add heavy image priors, such as Total Variation \cite{mahendran2015understanding} or Gaussian Blur \cite{yosinski2015understanding}, which produces naturally looking images.

The model inversion attacks were proposed, which adopts gradient-based inversion of the training classes to the task of face recognition \cite{fredrikson2015model}, leaking some representative images from training data, however it was shown that for deep convolutional neural networks it is notoriously difficult \cite{yang2019adversarial,hitaj2017deep}. This method also used denoising and sharpening filters as the prior.

Another category of reconstruction of image representations is the training-based inversion: an additional neural network is trained to map a feature vector into an image. The resulting neural network is similar to a decoder part of an auto-encoder network with the face identification encoder. To train a network, usually, L1 or L2 loss is used between original and reconstructed images. The results of this method were shown in \cite{dosovitskiy2016generating,dosovitskiy2016inverting,nash2018inverting}. Compared to gradient-based inversion, training-based inversion is only costly during training the inversion model, which is a one-time effort. Reconstruction from a given prediction requires just one single forward pass through the network.

Training-based methods to recover faces from the facial embeddings were found to produce interesting results in \cite{zhmoginov2016inverting,cole2017synthesizing,mai2018reconstruction,mohanty2007scores,mignon2013reconstructing}. In \cite{mignon2013reconstructing}, it was proposed to use the radial basis function regression to reconstruct faces from its signatures. In \cite{mohanty2007scores}, multidimensional scaling was used to construct a similarity matrix between faces and embeddings. It should be mentioned, both \cite{mignon2013reconstructing} and \cite{mohanty2007scores} were only tested for shallow neural networks. In \cite{zhmoginov2016inverting}, it was proposed to train a convolutional neural network that maps face embeddings to the photographs, however their method requires gradients of a face identification system. In \cite{cole2017synthesizing}, it was proposed to yield a reconstructed image from estimated face landmarks and textures, however high-quality face images are required for estimation. In \cite{mai2018reconstruction}, it was proposed to use the neighbourly de-convolutonal neural network to reconstruct recognizable faces, however this method requires input-output pairs for training process, and thus might be overfitted towards dataset or face identification model. To the best of our knowledge, no prior work on black-box zero-shot face reconstruction from identity features was presented before. To fill this gap, we propose our method. Since most of published results consider white-box setup, as direct competitor for our solution we see NBNet \cite{mai2018reconstruction}. Brief comparison of various methods is collected in Table \ref{comp_table}.

\subsection{Black-box mode, prior knowledge and number of queries}
In this paper, we consider a black-box attack procedure: we do not have access to the face recognition system and can only query it to get the output. In this setup, the number of queries is the main performance metric, along with the attack success rate. However, the number of queries is highly dependent on prior knowledge about the model. For adversarial examples this phenomenon is studied in \cite{ilyas2018prior}. Even models that are claimed to be fully black-box, such as NBNet \cite{mai2018reconstruction}, in fact exploit deep prior about the target model. They need to have a dataset from the same domain and the same alignment as it was for the target model, otherwise they cannot learn the function between a face image and an embedding since for the face-id network, this function is guaranteed to work well only for properly aligned images. In practice, we can have the model with a proprietary aligner and an unknown training dataset domain. One of the advantages of our method is that it is fully black-box and can work even in such a restrictive setup.

%-------------------------------------------------------------------------
\section{Gaussian sampling algorithm}
We designed an iterative algorithm for reconstructing a face from its embedding. The algorithm is a zero-order optimization in the linear space of 2D Gaussian functions. One step of our algorithm is the following: we sample a batch of random Gaussian blobs and add them to the current state image. Then the batch is put into black-box feature extractor, and loss function is evaluated across embeddings. Based on the evaluation, one image is selected and set as the current image. Such a procedure is similar to the random descent in the linear space of 2D Gaussian functions.
 
 \subsection{Choosing function-family for sampling}

 \begin{figure}[h]
\begin{center}
   \includegraphics[width=0.25\linewidth]{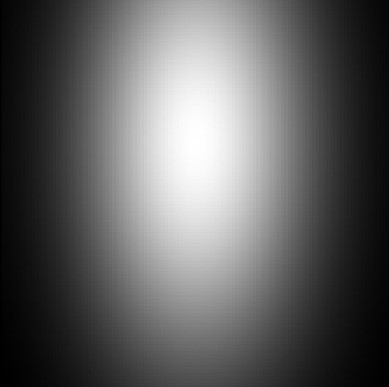}
\end{center}
   \caption{Gaussian blob with parameters: $x_0,y_0,\sigma_1,\sigma_2,A = 56, 72, 22, 42, 1$.}
\label{gauss1}
\end{figure}
 In our algorithm we sample Gaussian functions (Fig.~\ref{gauss1}):
$$
G(x,y) = A\cdot \exp{\frac{(x-x_0)^2}{2\sigma_1^2}}\exp{\frac{(y-y_0)^2}{2\sigma_2^2}}
$$
where,

\begin{tabular}{ll}
    $x,y$ - \text{pixel coordinates in the image,}\\
    $x_0,y_0$ - \text{coordinates of a center of gaussian,}\\
    $\sigma_1,\sigma_2$ - \text{vertical and horizontal standard deviations,}\\
    $A$ - \text{amplitude}\\
\end{tabular}
\hspace*{3em}\\

 Hypothetically, any function representing a basis in a 2D space can be chosen as a function for the sampling. We tried sines/cosines, polynomial functions, random noise, but only Gaussians-based approach works well. We suppose that the reasons are the following:
 \begin{enumerate}
    \item  Gaussian functions are semi-local, which means that the distortion of a picture is localized and hence more controllable. With even a small number of such functions, it is easier to fit many shapes.
    \item Low frequencies are dominant in Gaussian functions (if we restrict the interval of possible $\sigma$). We suppose this prevents overfitting of an attacked network and prevents generating of non-semantic high-frequency adversarial patterns.
\end{enumerate}

 We found that the restriction of the vertical symmetry on the family of sampling functions improves the speed of convergence and the quality of the final result, which makes sense as human faces are mostly symmetrical, and bringing this constraint to our algorithm reduces search space. We symmetrize sampled Gaussians by adding a vertically flipped copy: 
 $$
 G_{sym} = G + \text{flip}(G)
 $$
 
\begin{figure}[h]
\begin{center}
   \includegraphics[width=0.9\linewidth]{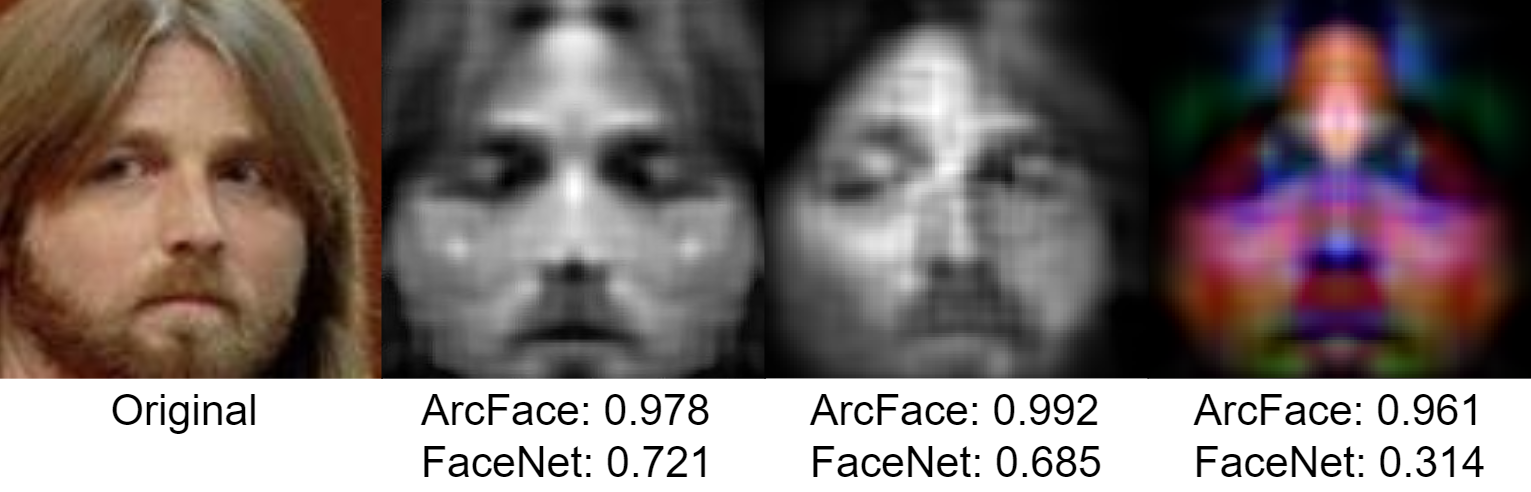}
\end{center}
   \caption{From left to right: original image, reconstructed from embedding in grayscale setting with symmetric constraint, without symmetric constraint, RGB with symmetric constraint and corresponding cosine similarities by attacked model (ArcFace) and independent (FaceNet).}
\label{color}
\end{figure}

\begin{figure}[h]
\begin{center}
   \includegraphics[width=0.8\linewidth]{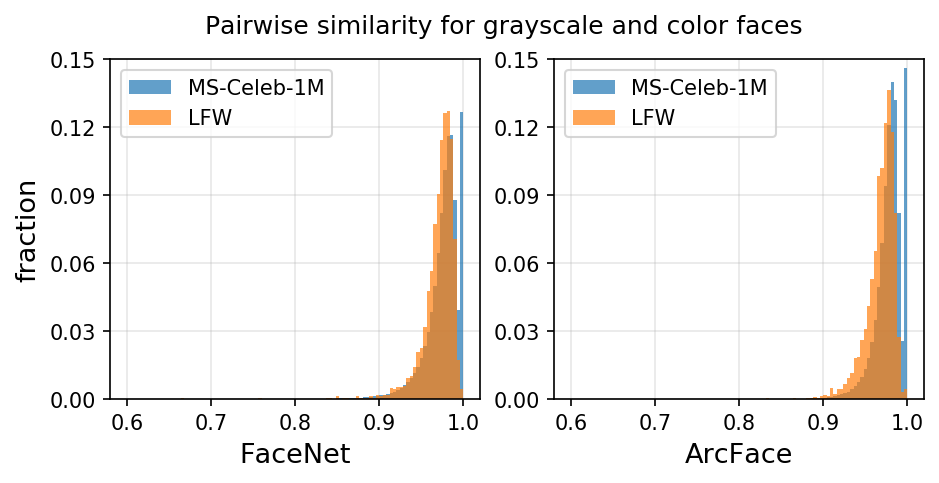}
\end{center}
   \caption{Pairwise ArcFace cosine similarity of images and their grayscale analogs from LFW and MS-celeb-1M datasets.}
\label{arcface_bw}   
\end{figure}

 To relax the problem further on, i.e. simplify the optimization process, we restore images in the grayscale colormap. In other words, the hypothesis is that embedding of deep face recognition systems is tolerant to color. 
 To verify the assumption, we set up two experiments for the most popular publicly available face recognition systems: ArcFace \cite{deng2019arcface} (model name "LResNet100E-IR, ArcFace@ms1m-refine-v2"\footnote{https://github.com/deepinsight/insightface/wiki/Model-Zoo}, accessed March 21, 2020) and FaceNet \cite{schroff2015facenet} (model name "20180402-114759"\footnote{https://github.com/davidsandberg/facenet}, accessed March 21, 2020). We checked pairwise similarity of RGB image and its grayscale copy. We perform this experiment with images from LFW \cite{huang2008labeled} and MS-Celeb-1M \cite{guo2016ms} (version named "MS1M-ArcFace"\footnote{https://github.com/deepinsight/insightface/wiki/Dataset-Zoo}, accessed March 21, 2020) datasets. It can be clearly seen on Fig~\ref{arcface_bw} that for the majority of images moving to the grayscale domain did not affect much corresponding embeddings. Anyway, we tried to reconstruct faces in the RGB domain, but obtained colors turned out to be far from natural regardless of the shapes being correct (Fig~\ref{color}). 
 
  So, our finding is that face embeddings are mostly not sensitive to color; therefore it is not possible to recover properly the color information of initial picture. Most importantly, it relaxes the problem significantly, allowing us to sample only grayscale Gaussian blobs. But, despite the fact that we reconstruct faces in a grayscale color space it is still possible to colorize it naturally later on with the use of dedicated colorization models \cite{zhang2016colorful}.

 \subsection{Loss function}
 A loss function is needed to choose the best sampled element from a batch. The suggested loss function depends on norms of embeddings (the target one and the embedding of a reconstructed image) and the cosine similarity between the target embedding and the embedding of a reconstructed image:
 $$
 L(y,y') =  \lambda\cdot(\|y\| - \|y'\|)^2 - s(y,y'),
 $$
%  where $s$ - cosine similarity, $\lambda=0.0025$
 where,\\
\hspace*{3em}
\begin{tabular}{ll}
    $s$ - cosine similarity function,\\
    $\|y\|$ - $\text{L}_2$ norm of the target embedding,\\
    $\|y'\|$ - $\text{L}_2$ norm of the embedding of \\
    a reconstructed image,\\
    $\lambda=0.0025$, empirically found hyperparameter\\
\end{tabular}
\hspace*{3em}\\
 
 \subsection{Initialization}
\begin{figure}[h]
\begin{center}
   \includegraphics[width=0.6\linewidth]{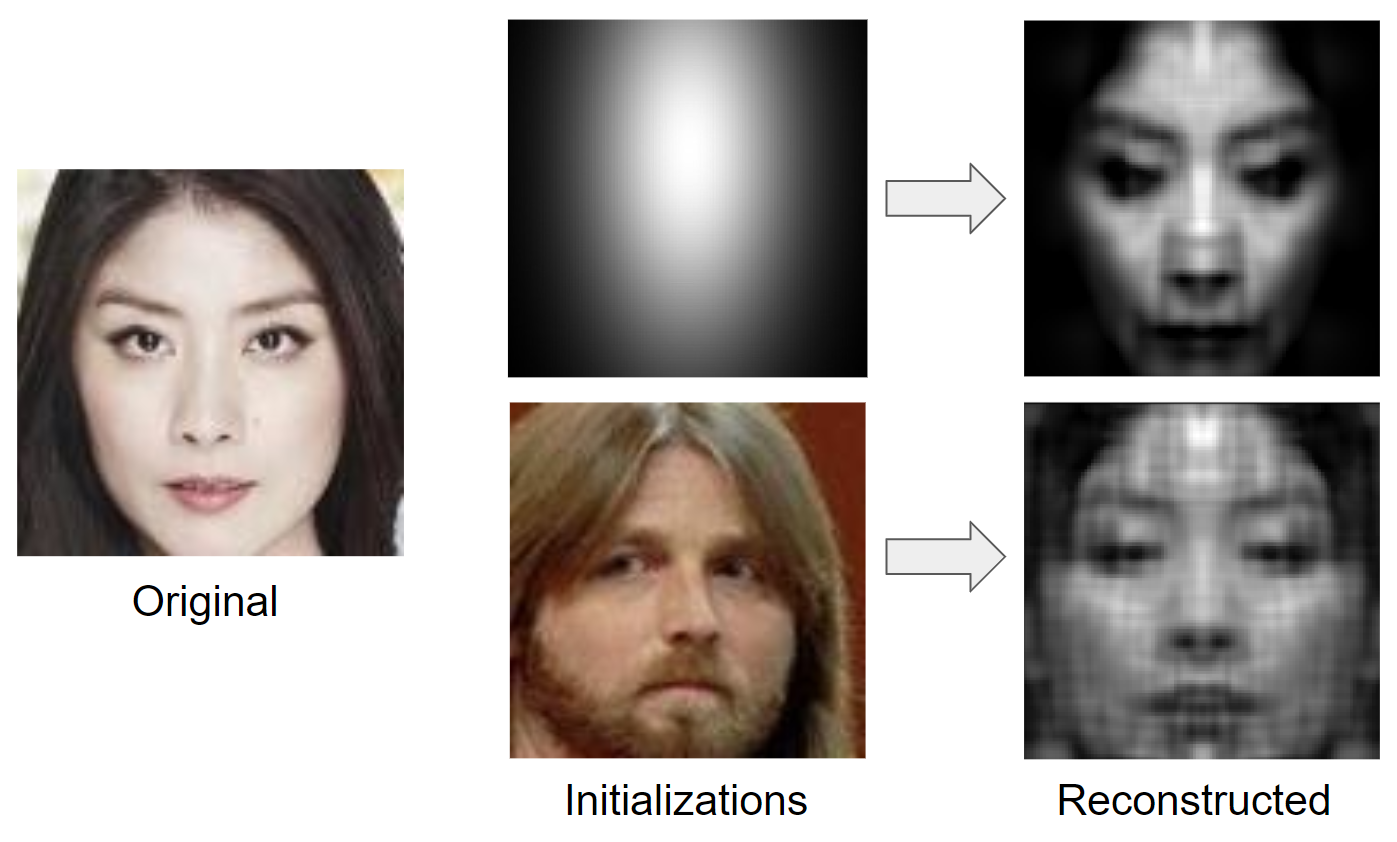}
\end{center}
   \caption{From left to right: original image, two types initializations: an optimal Gaussian blob and a random face, corresponding reconstructed images.}
\label{initialization}
\end{figure}

We found that proper initialization of the algorithm has crucial importance. Without an initialization, the algorithm most likely would not converge to a face. We tried two variants of initialization (Fig~\ref{initialization}):
 \begin{enumerate}
  \item Initialize with a face. This kind of initialization uses a predefined image with a face. Such initialization works good and even let us not use  norm of an embedding (works just with cosine similarity between embeddings as a loss function). But it has a strong disadvantage as the reconstructed face is "fitted" into initialization face: reconstructed face looks similar to a target person (facial traits), while it has the shape of initialization face. Thus, we did not use this method for further experiments.
  \item Initialize with the optimal Gaussian blob (optimal in terms of cosine similarity between the target embedding and the Gaussian blob). We constructed a set of 4480 vertically symmetric Gaussians, which are similar to the shape of natural faces. Then we search for the best one for a given embedding by comparing cosine similarities. This kind of initialization requires adding the norm of an embedding to the loss function as, without it, it would not converge to a face-like picture. 
\end{enumerate}
For both initializations, we fade-out the initialization part of the reconstructed image at every iteration by multiplying it by fade-out coefficient 0.99.

\subsection{Validation using independent critic}
While comparing the results of different variations of the algorithm, we faced a problem of the objective evaluation of the quality of a reconstructed face. In \cite{mai2018reconstruction} and \cite{cole2017synthesizing} the cosine similarity between embeddings of the target image and the reconstructed one was used as a criterion of quality, but they used the same network for evaluation as for reconstruction attack, that, we think, might cause some problems as the reconstructed face might have high similarity with the true image but does not look like the same person and even does not look face-like --- so it is a kind of "adversarial face" which has high similarity but looks wrong. This happens as the network used for evaluation is the same as used in an algorithm, and it is a "dependent critic". What is more important, because of the specificity of our algorithm (minimizing loss), the reconstructed faces always have cosine similarities higher than 0.9 when attacking the same network as for similarities computation. To quantify the quality of a reconstructed image in a robust way, we suggest using another network with different architecture compared to the attacked one as an "independent critic". Another solution is to use the human evaluation (like Mean Opinion Score), but as human opinion varies -- some statistics are needed to quantify the quality of compared images.

We used FaceNet trained on VGGFace2 \cite{cao2018vggface2} as an independent critic. We provide results for both metrics ---  "dependent critic" and "independent critic".

\subsection{Algorithm}
\begin{figure}
\begin{center}
   \includegraphics[width=\linewidth]{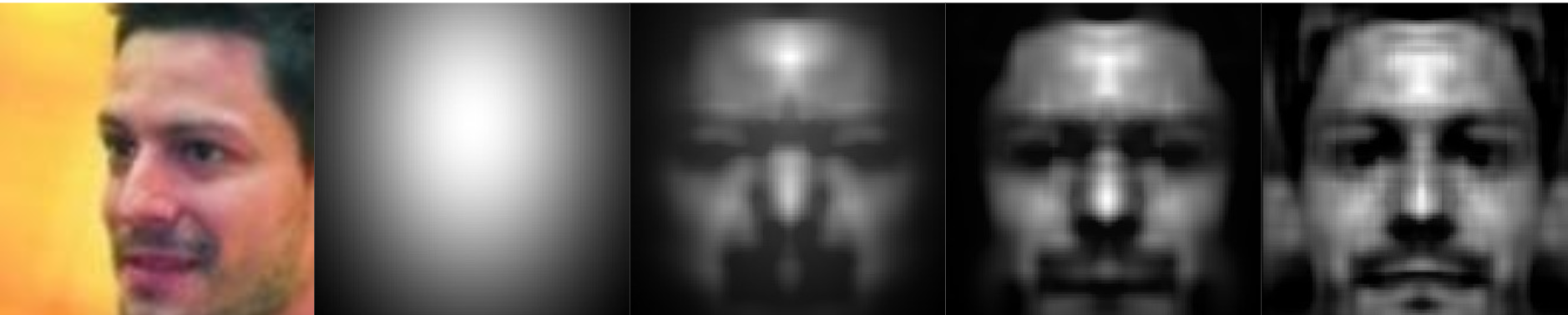}
\end{center}
   \caption{Iterations of Gaussian sampling algorithm. From left to right: original image, initialization, 30k queries, 60k queries, 300k queries.}
\label{process}
\end{figure}

\begin{figure}[h]
\begin{center}
   \includegraphics[width=0.7\linewidth]{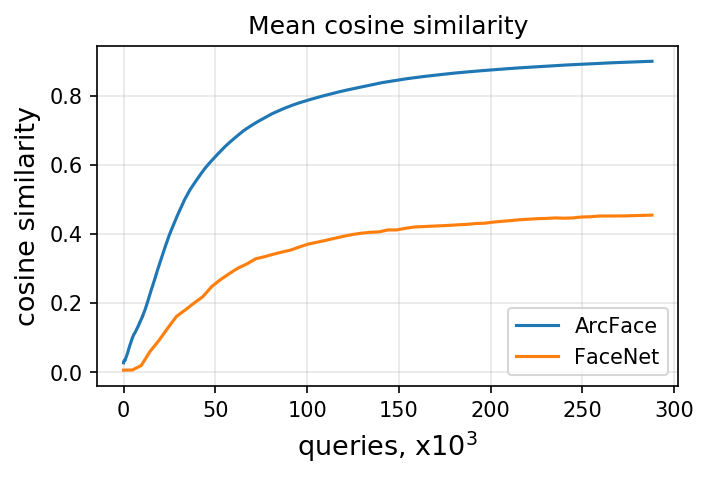}
\end{center}
   \caption{Mean cosine similarity between target embedding and embedding of reconstructed image for filtered LFW subset for our algorithm.}
\label{mean_cosine}
\end{figure}

The algorithm is a zero-order optimization in the space of 2D Gaussian functions. At every step the best one Gaussian function from a batch is chosen in terms of objective function and added to current reconstruction image.
The formal description of an algorithm is presented using Algorithm \ref{algorithm}. An example of the reconstruction process is presented in Fig.~\ref{process}. The mean cosine similarity dynamics while doing queries is presented on Fig.~\ref{mean_cosine}.

\begin{algorithm}
\caption{Face recovery algorithm}\label{alg}
\textbf{INPUT:} target face embedding $y$, black-box model $M$, loss function $L$, $N_{queries}$
\begin{algorithmic}[1]
\State $X\leftarrow 0$
\State Initialize $G_0$
\For{$i\leftarrow 0$ to $N_{queries}$}:
\State Allocate image batch \textbf{X}
\State Sample batch $\textbf{G}$ of random gaussians
\State $\textbf{X}_j = X + G_0 + \textbf{G}_j$
\State $\textbf{y}' = M(\textbf{X})$
\State $\text{ind} = \text{argmin}\biggl( L(\textbf{y}_i',y)\biggr)$
\State $X \leftarrow X + \textbf{G}_{\text{ind}}$
\State $G_0 \leftarrow 0.99\cdot G_0$
\State  $i \leftarrow i + \text{batchsize}$
\EndFor
\State $X \leftarrow X + G_0$
\end{algorithmic}
\label{algorithm}
\textbf{OUTPUT:} reconstructed face $X$
\end{algorithm}

%-------------------------------------------------------------------------
\section{Experiments}

\subsection{Baseline reproduce}
We use the original NBNet source code (author's git repository\footnote{https://github.com/csgcmai/NBNet}, accessed March 21, 2020) and trained it on MS1M-ArcFace dataset. Retrain is needed since the original model is trained with different alignment and worked poorly with photos aligned for ArcFace (by MTCNN \cite{zhang2016joint}). In the original paper, it was trained on the DCGAN output, since there were no sufficient datasets at the time of publication. However, modern datasets are much larger than the number of queries needed for NBNet (MS1M-ArcFace contains 5.8M images). We followed the original paper training procedure as far as it was possible. The model was trained with MSE loss at the first stage, then the perceptual loss was added at the second stage. The MSE loss stage took 160K$\times$64 queries, then the loss stopped to decrease. The perceptual loss stage took 100K iterations, as in the original paper. 

\subsection{Face reconstruction}

\begin{figure*}[t]
\begin{center}
   \includegraphics[width=0.9\linewidth]{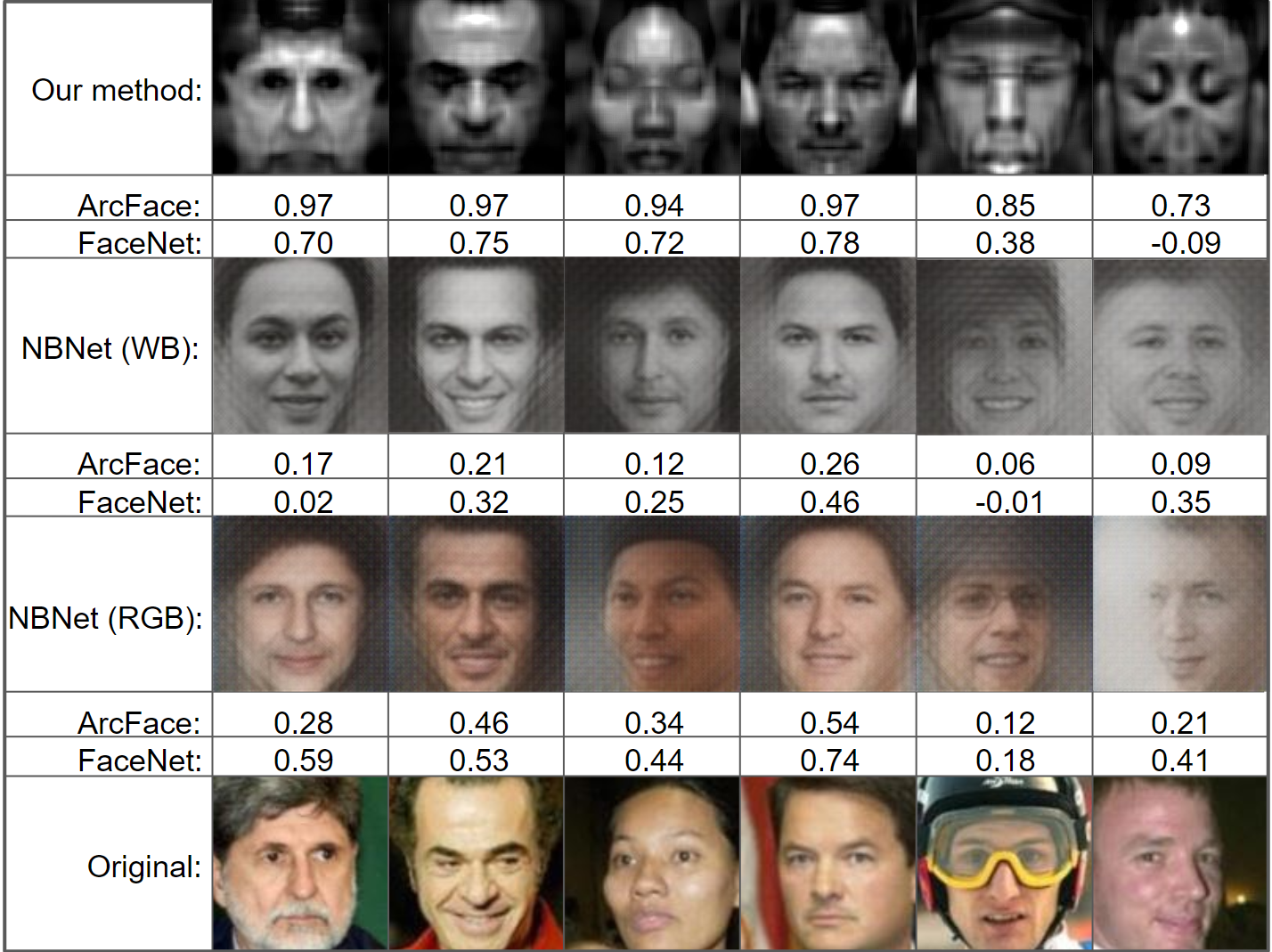}
\end{center}
   \caption{Examples of recovered images from LFW dataset and the corresponding cosine similarities by ArcFace and FaceNet.}
\label{lfw_examples}
\end{figure*}

\begin{table}
\begin{center}
\begin{tabular}{|l|c|c|c|}
\hline
Method & ArcFace & FaceNet & \# of queries\\
\hline\hline
(Ours) Symmetric gauss, LFW (wb)& $\textbf{0.90}$ & {$\textbf{0.45}$} & $\textbf{300}$k\\\hline
(Ours) Asymmetric gauss, LFW  (wb)& $0.85$ & $0.43$ &$400$k\\\hline
NBNet, LFW (RGB)&$0.25$& $0.34$& $3\text{M}$\\\hline
NBNet, LFW (wb)&$0.19$& $0.27$& $3\text{M}$\\\hline\hline

(Ours) Symmetric gauss, MS1M-ArcFace (wb) & $\textbf{0.89}$ & $\textbf{0.42}$ & $\textbf{300}$k \\\hline
NBNet, MS1M-ArcFace (RGB) &$0.26$& $0.38$&  $3\text{M}$\\\hline
NBNet, MS1M-ArcFace (wb) &$0.20$& $0.32$&  $3\text{M}$\\\hline
\end{tabular}
\end{center}
\caption{Average cosine similarity by ArcFace and FaceNet (independent critic) between embedding of a reconstructed image and embedding of target image  for subsets of 100 images from LFW and MS1M-ArcFace and corresponding number of queries.}
\label{results1}
\end{table}

To evaluate our method we considered two main setups:

\begin{enumerate}
    \item Reconstruction of faces from a MS-Celeb-1M dataset the attacked network (ArcFace) is trained on. We used a random  subset of 100 faces of different persons (identities), aligned with MTCNN;
    
    \item Reconstruction of faces that are not presented in the training dataset. We selected a subset of 100 unique faces of different persons (identities) from LFW that are not presented in MS1M-ArcFace: we checked each identity in LFW with all identities given in MS1M-ArcFace and left only ones for which cosine similariy was below $0.4$. All images are aligned with MTCNN too.
    
\end{enumerate}
Two sets of images are reconstructed: one with NBNet and another one with the proposed approach. The obtained faces are then fed to ArcFace and FaceNet to check the cosine similarity distribution. These setups allow the performance of given methods to be evaluated in two scenarios:  

\begin{enumerate}
    \item The network has already seen the photo, so it would ease the problem;
    \item The network has never seen the photo to be restored.
\end{enumerate}

\begin{figure}[h]
\centering
\begin{subfigure}{.5\textwidth}
  \centering
  \includegraphics[width=\linewidth]{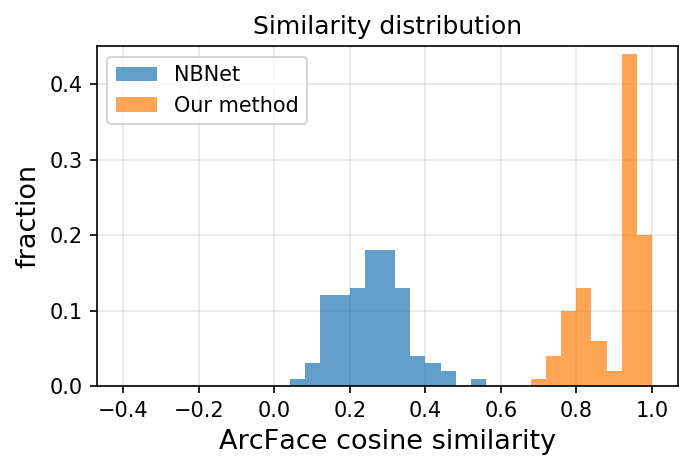}
  \caption{ArcFace}
  \label{fig:sub1}
\end{subfigure}%
\begin{subfigure}{.5\textwidth}
  \centering
  \includegraphics[width=\linewidth]{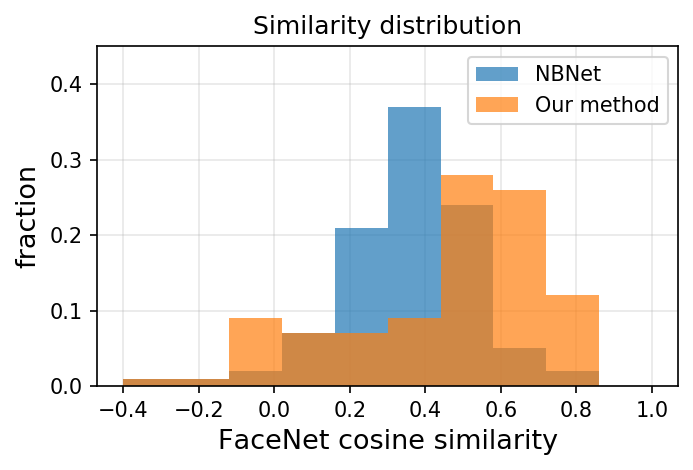}
  \caption{FaceNet}
  \label{fig:sub2}
\end{subfigure}
\caption{cosine similarity distribution for reconstructed faces and their true embeddings for subset of 100 unique identities from LFW.}
\label{fig:test}
\end{figure}

In order to provide honest comparison, we trained NBNet with the same hyperparameters on grayscale version of dataset. Since our method restores a grayscale image, we thought that NBNet could also benefit from the color reduction.
The results for the first setup are illustrated in Fig.~\ref{fig:test}. The first figure presents cosine similarity distribution for ArcFace network, where our method shows the superior performance. However, since ArcFace is the attacked model, such results might be caused by overfitting. In order to avoid this, we also checked results with FaceNet. Facenet results also shows superior performance: the distribution of faces generated by the proposed method is shifted towards a higher similarity range compared to NBNet. Also it is shown that, in fact, grayscale train degrades NBNet performance.

We also checked reconstruction for the symmetric and asymmetric modes for LFW and MS1M-Arcface datasets.  Since the symmetric mode reduces the complexity of an optimization process, it should show a superior performance taking less number of queries compared to the asymmetric mode: which is confirmed experimentally, and results are shown in Table~\ref{results1} for both datasets. 

The reconstruction process is shown in Fig.~\ref{process} with validation on FaceNet. The obtained faces are given on Fig.~\ref{lfw_examples}. We observed interesting behavior in the reconstruction process: faces with high validation similarity always have high-quality attributes while low similarity faces have a rather unnatural look (can be seen on the last column in Fig.~\ref{lfw_examples}). This is completely different from what happens with the face reconstruction by NBNet, where faces are always good to look at, and the quality does not correlate much with the cosine similarity. We attribute this problem to the NBNet training procedure, which optimizes MSE loss, which depends on insignificant features such as skin tone, while important features (eyebrows, nose form and so on) impact slightly.

\section{Conclusion \& Future Work}
We demonstrate that it is possible to recover recognizable faces from deep feature vectors of a face-recognition model in a black-box mode with no prior knowledge. The proposed method outperforms current solutions not only in terms of the average cosine similarity of embeddings produced by the attacked model but in terms of average cosine similarity given by an independent critic. Moreover, the proposed method requires a significantly smaller number of queries compared to previous solutions and does not need prior information such as proper training dataset, in other words -- our algorithm works in a zero-shot mode and hence does not need to know how faces look like to recover them. As a future work, we see an investigation of poorly reconstructed faces and further minimization of the number of queries.

\clearpage
% ---- Bibliography ----
%
% BibTeX users should specify bibliography style 'splncs04'.
% References will then be sorted and formatted in the correct style.
%
\bibliographystyle{splncs04}

\bibliography{egbib.bib}

\end{document}